\documentclass[runningheads]{llncs}
\usepackage{graphicx}
\usepackage{amsmath, amssymb}
\usepackage{subcaption}
\usepackage{url}
\usepackage{xurl}

\usepackage[normalem]{ulem}
\usepackage{comment}
\usepackage{multirow}
\usepackage[table,xcdraw]{xcolor}
\usepackage{booktabs, cellspace}
\PassOptionsToPackage{hyphens}{url}\usepackage{hyperref}

\usepackage{caption} 
\begin{document}
\title{LLCoach: Generating Robot Soccer Plans using Multi-Role Large Language Models}
%
%
\author{
M. Brienza \inst{1} 
\orcidID{0009-0000-1549-9500}
\and
E. Musumeci \orcidID{0009-0004-2359-5032} \inst{1}
\and
V. Suriani \inst{2}  
\orcidID{0000-0003-1199-8358}
\and
D. Affinita \inst{1}
\and
\\A. Pennisi \inst{3}\orcidID{0000-0002-9081-0765}
\and
D. Nardi\inst{1}\orcidID{0000-0001-6606-200X}
\and
\\D. D. Bloisi \inst{3}\orcidID{0000-0003-0339-8651}
}
\authorrunning{Brienza, Musumeci et al.}
\titlerunning{LLCoach: Generating Robot Soccer Plans using Multi-Role LLMs}
%
\institute{Dept. of Computer, Control, and Management Engineering\\ Sapienza University of Rome, Rome (Italy),
    \email{\{lastname\}@diag.uniroma1.it} \and
School of Engineering, University of Basilicata, Potenza (Italy),
\email{vincenzo.suriani@unibas.it} \and
Dept. of International Humanities and Social Sciences, 
International University of Rome, Rome (Italy),
\email{domenico.bloisi@unint.eu}
}
\maketitle              
\begingroup\renewcommand\thefootnote{\textsection}
\endgroup
\begin{abstract}
The deployment of robots into human scenarios necessitates advanced planning strategies, particularly when we ask robots to operate in dynamic, unstructured environments. RoboCup offers the chance to deploy robots in one of those scenarios, a human-shaped game represented by a soccer match. In such scenarios, robots must operate using predefined behaviors that can fail in unpredictable conditions.
This paper introduces a novel application of Large Language Models (LLMs) to address the challenge of generating actionable plans in such settings, specifically within the context of the RoboCup Standard Platform League (SPL) competitions where robots are required to autonomously execute soccer strategies that emerge from the interactions of individual agents.
In particular, we propose a multi-role approach leveraging the capabilities of LLMs to generate and refine plans for a robotic soccer team. 
The potential of the proposed method is demonstrated through an experimental evaluation, 
carried out simulating multiple matches where robots with AI-generated plans play against robots running human-built code.

\end{abstract}

\section{Introduction}

\begin{figure*}[h!]
  \centering
  \includegraphics[width=0.8\textwidth]{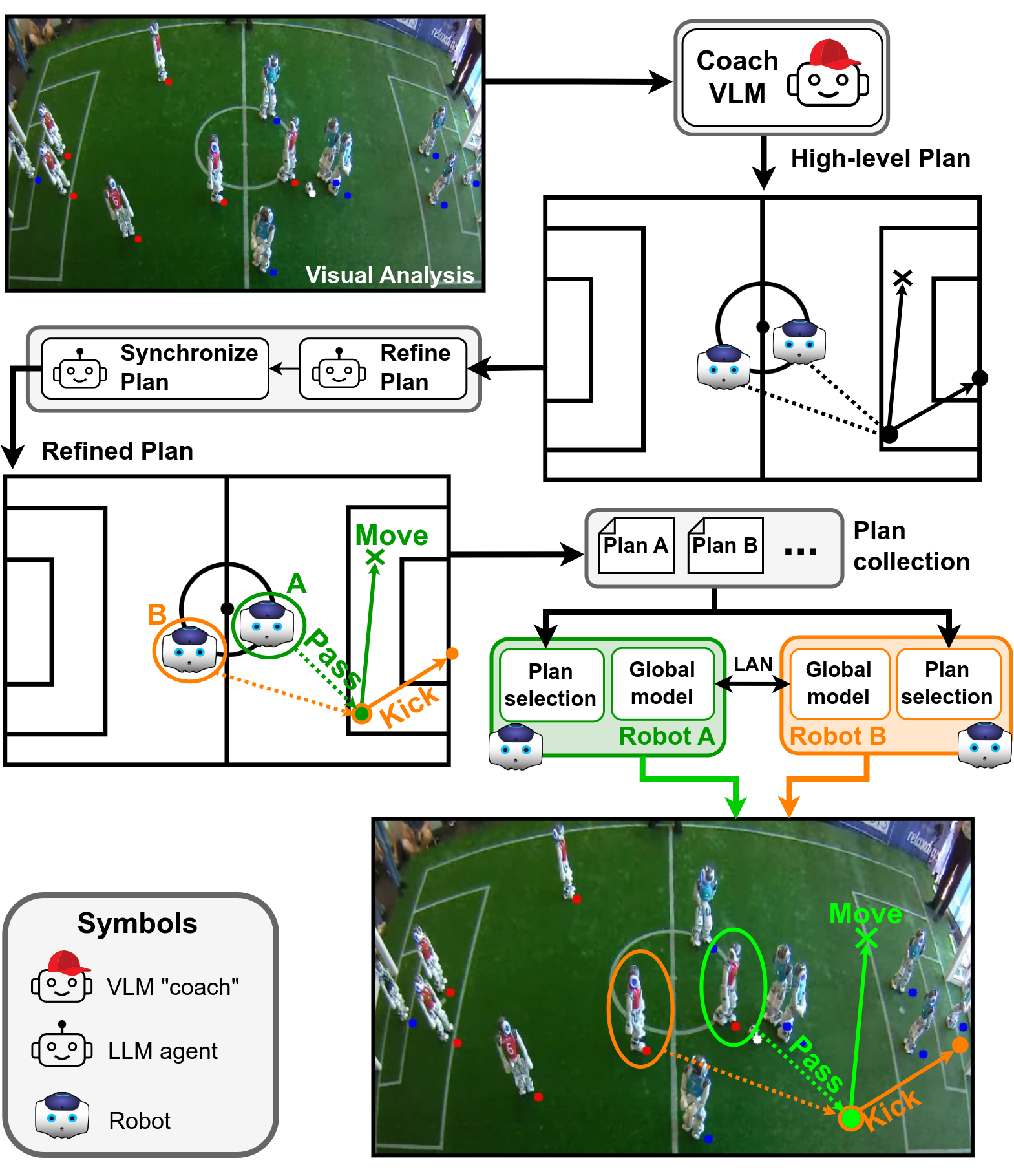}
  \caption{The LLCoach architecture includes high-level plan generation by the VLM coach, which is then refined to a low-level plan, executable in robot matches.}
  \label{fig:intro}
  \vspace{-0.5cm}
\end{figure*}

Recent advancements in Language Modeling, particularly the proliferation of technologies associated with Large Language Models (LLMs), have unlocked numerous ways to enhance embodied AI systems with common sense knowledge \cite{NEURIPS2023_65a39213}. The expansive scope of LLMs serves as a vast repository of information, facilitating multi-modal interactions with textual and visual data.

When appropriately queried, LLMs provide access to semantic cues that are crucial for bridging the conceptual gap between abstract planning domains and intricate and unpredictable real-world environments \cite{song2023llmplanner}. Consequently, they are increasingly integrated into embodied AI systems \cite{10373065}.

In this work, we present a first attempt to use generative AI in the context of robot soccer for creating successful game plans using an approach to leverage the LLMs' chain of thought and generate game tactics by impersonating a coach.
In particular, we present a multi-role pipeline, featuring four stages in which the capabilities of a foundation model are exploited to generate and refine a plan for a soccer robot (see Fig. \ref{fig:intro}).
Robot soccer offers an ideal environment for planning within a multi-robot, highly dynamic environment \cite{best2023}.
Here, we focus on the RoboCup Standard Platform League (SPL), which involves teams of Aldebaran NAO V6 humanoid robots competing autonomously in soccer matches. Participating teams are tasked with executing multi-agent strategies aimed at securing victory in the matches.
It is worth noting that, in SPL matches, the applicability of commonsense knowledge is constrained, given that a detailed understanding of game rules and the capabilities of robots during soccer matches is often specialized knowledge. Access to such information, including the actions available to each agent, is typically facilitated through a Retrieval-Augmented Generation (RAG) system \cite{balaguer2024rag}.
The contribution of this work is two-fold: 
\begin{enumerate}
    \item A methodology for extracting high-level strategies from videos of RoboCup soccer matches.
    \item A pipeline for creating detailed, multi-agent plans that can be directly executed by a NAO robot using its available set of basic actions. 
\end{enumerate}

Experimental results are obtained in a simulated environment by performing multiple matches where the robots with the AI-generated plans play against the robots running the code\footnote{\url{https://github.com/SPQRTeam/spqr2023}} used by SPQR Team in RoboCup 2023.
The code and the additional material mentioned in this paper are publicly available at \url{https://sites.google.com/diag.uniroma1.it/llcoach/}.

The rest of the paper is organized as follows. Section \ref{sec:relwork} presents a brief overview of related work, while
Section \ref{sec:method} showcases the methodology and the proposed architecture. Experimental results are shown in Section \ref{sec:ExperimentalResults}. Finally, Section \ref{sec:conclusion} draws the conclusions and future directions.

\section{Related Work}
\label{sec:relwork}


Before the advent of the LLMs, few approaches have been proposed for coaching teams of robots. The first attempt is represented by a language to coach a RoboCup team\cite{reis2002coach}, where COACH UNILANG is presented for Simulated 3D League. More recently, in SPL, \cite{musumeciCoach} presented an approach to conditioning the robot's behavior before or during the time of the match. 

More recently, LLMs have been used as planners for both structured \cite{silver2023generalized} and unstructured environments \cite{huang2022language}.
However, when using LLMs as planners, especially for embodied AI (where visual cues must be conveyed through text), there is the possibility of incurring hallucinations, causing skewed or imprecise results.
Hallucination is a phenomenon typically observed when the provided textual inputs are too structured and long. For such a reason, it proves beneficial to provide examples of desired output and to define roles for the LLM in a request, by instructing the model on its role and providing constraints on how to perform its task. Demanding to each role a very specific task in a pipeline composed of multiple steps allows to obtain a better final result that would be too complex for the model with a single, lengthy, and complex query.
Thanks to the introduction of VLMs \cite{openai2023gpt4v}, LLMs can get both textual and visual inputs reducing the length of the context that must be provided textually in scenarios where visual cues are essential to understand the requested task. Such an approach alleviates the issue of hallucination in applications of embodied AI.
However, due to their considerable size in terms of parameters and the consequent computational demands for a single training or inference step, LLMs are less likely to be employed in embedded systems. Moreover, fine-tuning them is an exceedingly resource-intensive endeavor, seldom justifiable for enhancing the accuracy of tasks reliant on specialized commonsense knowledge. 

To better fit unknown domains, Retrieval-Augmented Generation (RAG) is implemented within intricate pipelines. Here, segments of potentially lengthy documents are indexed based on their semantics in an embedding space and stored within a Vector Index. 
This approach facilitates enhanced and expedited customization of LLM query results compared to traditional fine-tuning methods. Moreover, additional pieces of information can be seamlessly integrated into the supplementary knowledge base by appending them to the Semantic Database. This capability enables, for instance, the generation of documents from semantically akin templates \cite{musumeci2024llm}.

RAG-based systems offer interesting applications to robotics. In \cite{yuan2024rag}, RAG-Driver is presented as a novel multimodal RAG LLM for high-performance, explainable, and generalizable autonomous driving. 
RAGs in robotics offer the possibility to enrich the available commonsense knowledge with more specialized knowledge about the environment or the robotic agent's capabilities. 



\section{Proposed Approach}
\label{sec:method}

\begin{figure}[h]
  \centering
  \includegraphics[width=0.95\textwidth]{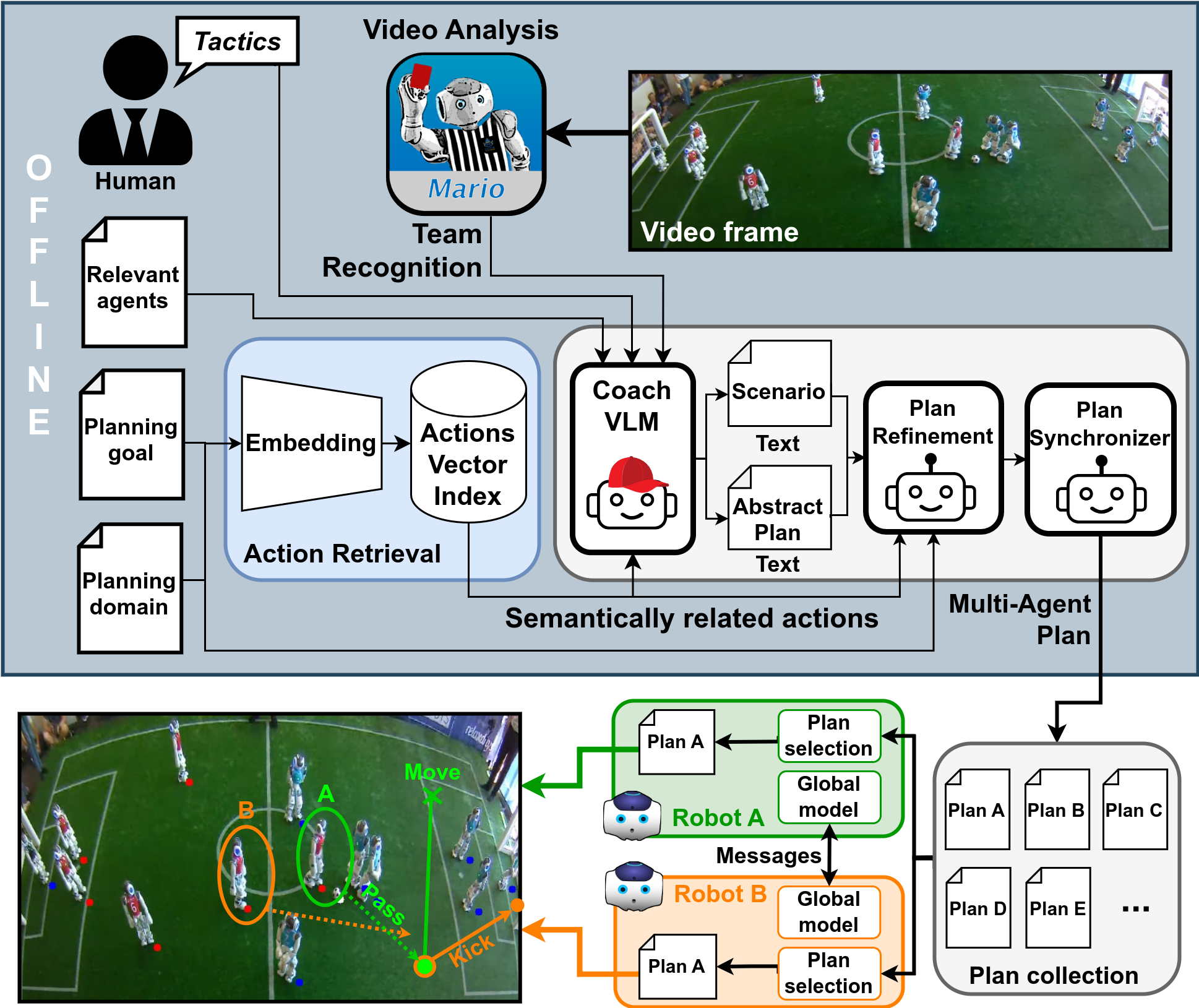}
  \caption{The pipeline is subdivided into an offline component and an online one, executed in real-time. The offline component collects plans by passing video frames to the coach VLM, and refines them using a multi-role LLM pipeline, using only actions obtained by RAG. The online component retrieves and executes the most fitting plan according to the world model, shared between robots.}
  \label{fig:pipeline}
  \vspace{-0.5cm}
\end{figure}

Fig. \ref{fig:pipeline} shows our multi-role LLM-based planning sequential pipeline, featuring four steps, namely \emph{Action Retrieval}, \emph{Coach VLM}, \emph{Plan Refinement}, and \emph{Plan Synchronizer}. 

\textbf{Input.} Although commonsense knowledge is usually enough for most generic tasks, in the case of robot soccer, specific knowledge must be provided. Thus, the prompts for the LLM-based generation steps along the pipeline are constructed from case-specific information.
In particular, case-specific knowledge includes:
\begin{itemize}
    \item The \textit{domain}, containing a natural language description of the application (a soccer match between autonomous robots), the definition of tokens representing the locations of waypoints, described in natural language as a discrete distribution of several relevant locations. This allows managing the complexity of planning in a continuous coordinate space by delegating the grounding of waypoint locations (into actual spatial coordinates) to the robot control framework, where a world model will be available.
    \item A \textit{planning goal}: "\textit{The own team should score a goal in the opponent's goal.}"
    \item A description of robot team formations and agent capabilities in natural language, containing tokens associated with the actions they can perform.
\end{itemize}

\textit{Actions} are described using natural language, following a structure similar to STRIPS actions (similar to Planning Domain Definition Language \cite{pddl}) where tokens used from the domain and agent descriptions are used to give the LLM a better understanding of their pre-conditions and effects. 

To prevent hallucination, the generation process is subdivided into several subsequent generation steps, centered on the interaction with an LLM or a VLM, each with a properly engineered prompt to minimize its size and increase the quality of the information provided to the model.

\subsection{Action Retrieval}

To minimize the size of the prompts sent to the LLM or VLM modules requiring an understanding of agent actions, in an attempt to prevent hallucination, actions are stored in a Semantic Database, where each data node corresponds to the vector embedding of their natural language description. 
The semantic database leverages a Vector Index for efficient storage and retrieval of action definitions.

The Action Retrieval module retrieves actions from the database based on their semantic similarity with the input plan. A \emph{Sentence Transformer}, specifically the MPNet model \cite{song2020mpnet}, is employed to compute this semantic similarity. This embedding space is crucial for both populating the database and performing retrieval operations. 
Actions stored in the database follow a STRIPS-like template:
\begin{small}
\begin{verbatim}
ACTION_ID: the name of the action
DESCRIPTION: a high-level description that explains what
the action does.
ARGS: ARG_NAME : ARG_VALUE specifies action arguments.
PRECONDITIONS: conditions that must be satisfied to execute the action.
EFFECTS: the expected result
\end{verbatim}
\end{small}
During retrieval, the planning domain and goal are also embedded within this same space, so that only relevant actions may be extracted using a k-nearest neighbor search based on cosine similarity, measuring the alignment between the required task and the actions stored in the database. This enables the extraction of pertinent actions only, minimizing the prompt length for subsequent pipeline steps and improving its quality.


\subsection{Coach VLM}
\subsubsection{Visual Analysis.}

To acquire the necessary frame information for the coach's utilization of the VLM to generate the desired output, we leverage MARIO \cite{bloisi2022mario}, a tool specialized in soccer match video analysis.


MARIO extracts a representation of the soccer game using its tracking capabilities and converts it into a 2D plan view applying a homography geometrical transformation. 
This representation is useful
to characterize video frames in a way that a VLM can easily understand. In particular, each robot is marked with a dot representing the color of the robot's jersey, which indicates the team color (see the bottom-left of Fig. \ref{fig:pipeline}).

\begin{figure}[t]
  \centering
  \begin{subfigure}[b]{0.46\textwidth}
    \centering
    \includegraphics[width=\textwidth]{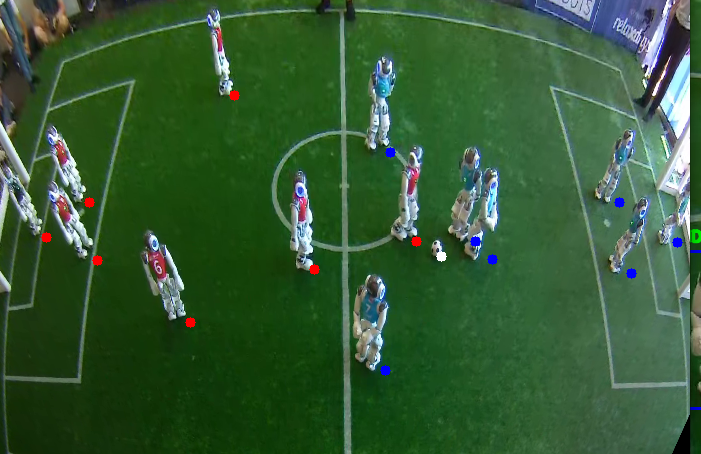}
    \caption{}
  \end{subfigure}
  \hfill
  \begin{subfigure}[b]{0.51\textwidth}
    \centering
    \includegraphics[width=\textwidth]{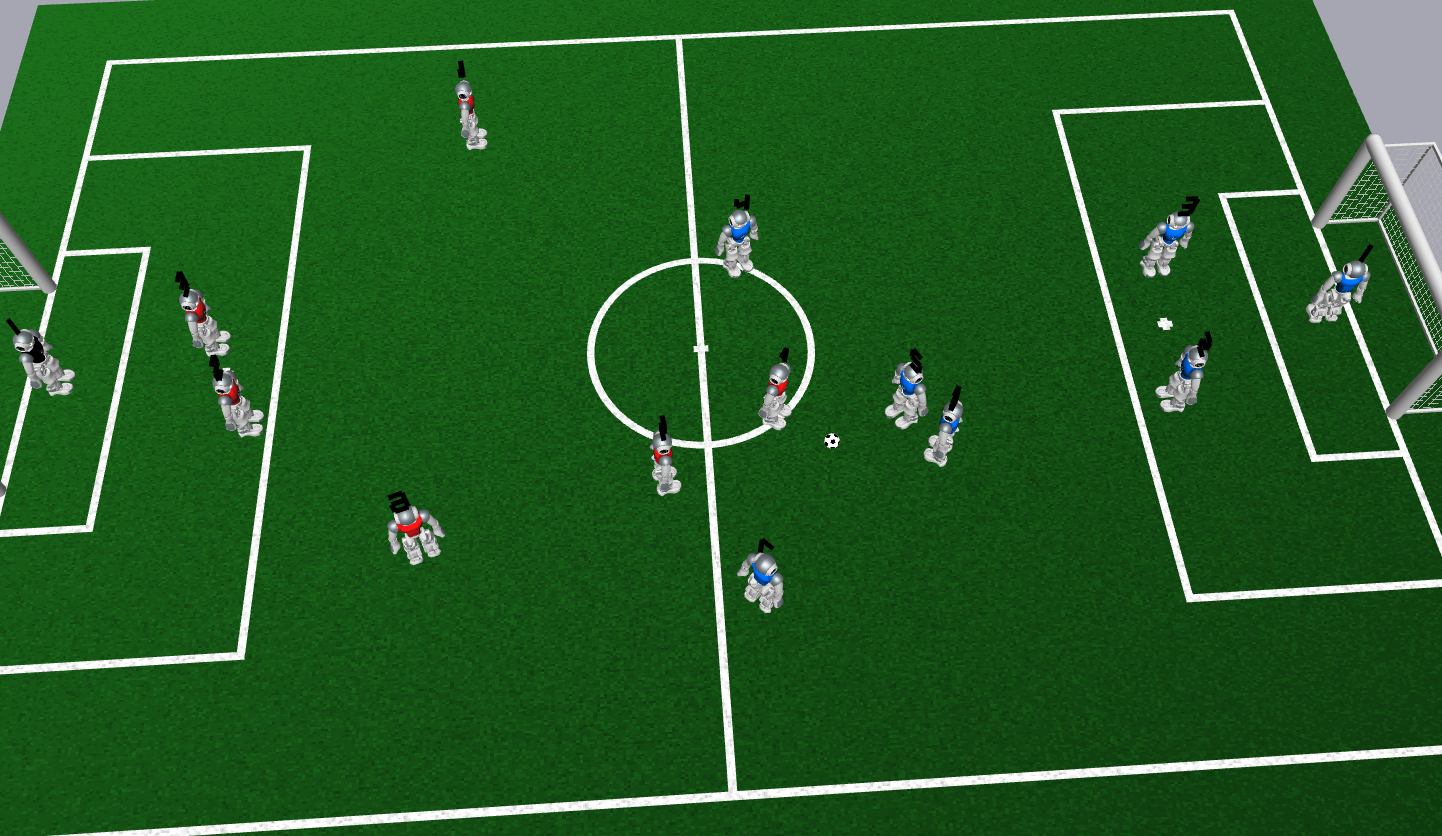}
    \caption{}
  \end{subfigure}
  \caption{MARIO visual tool, transforming robot pose (original (a), simulation (b)).}
  \label{fig:result_action}
  \vspace{-0.5cm}
  
\end{figure}

\subsubsection{VLM module.} The \textit{coach} module exploits the capabilities of GPT4-Vision \cite{openai2023gpt4v}, a recent Visual Language Model (VLM). A video frame resulting from the visual analysis of a relevant action, like the one in Fig. \ref{fig:result_action}, is provided to the Visual Language Model, along with a textual prompt. 
While the video frame provides a detailed view of the current configuration of agents and obstacles on the field, the textual prompt is used to instruct the VLM on its task and constraints. In particular, the coach is instructed to perform two tasks: 
\begin{enumerate}
    \item Generating a schematic description of the video frame.
    \item Generating a high-level plan in natural language.
\end{enumerate}
The prompt for the first task is based on the following template.
\begin{small}
\begin{verbatim}
As a coach assisting the red team, provide a precise summary using the 
formatting below. Each player's position should be related to designated 
waypoints on the field.
SCENARIO:
[ROLE_OWN_TEAM] is at [WAYPOINT]
[ROLE_OPPONENT_TEAM] is at [WAYPOINT]
...
BALL is at [WAYPOINT]    
\end{verbatim}
\end{small}
Given the proven capabilities of LLMs as few-shot learners \cite{song2023llmplanner}, the prompt is enriched by an example of an ideal output structure.
Then, a description of the robot roles in the multi-agent team is added, followed by a list of the available movement waypoints, with a semantic description for each one. Upper-case words are used in an attempt to increase attention over specific tokens.

The second task consists of generating a high-level plan in natural language, describing the optimal strategy to achieve the planning goal, given the same inputs as the first task. The prompt follows the template:
\begin{small}
\begin{verbatim}
After describing the SCENARIO, provide coaching advice.
Your suggestions should utilize the following actions: [ACTIONS]
Recommend player movements to achieve the objective [PLANNING_GOAL]
Your attitude is to perform the following tactics [TACTICS]
\end{verbatim}
\end{small}
Here:
\begin{itemize}
    \item \textit{[ACTIONS]} is replaced with actions extracted at the retrieval step.
    \item \textit{[PLANNING\_GOAL]} is the original planning goal for the current task, which consists simply in "\textit{The own team should score a goal in the opponent's goal.}"
    \item \textit{[TACTICS]} allow additional customization of the planning goal (which is instead fixed).
\end{itemize}
In addition to the task, several constraints over the desired output are added:
\begin{small}
\begin{verbatim}
Respect the sequential execution of actions,considering that the previous 
action changes the game situation. Constraints on passing: a robot 
cannot pass or kick the ball if it has passed it before, receive 
the ball only if you are at the target location otherwise, consider
robot movement actions.
\end{verbatim}
\end{small}
Then an example of ideal output is provided, to enforce a specific structure:
\begin{small}
\begin{verbatim}
COACH ADVICE:
[give the actions to be performed avoiding conditionals 
but describing the precise steps to be performed in sequence.]
\end{verbatim}
\end{small}
The output of the VLM is the concatenation of the responses to the two tasks:
\begin{enumerate}
    \item A structured description of the scenario represented in the input video frame.
    \item A multi-agent high-level plan, featuring all the relevant agents detected in the provided snapshots, that represents the suggested strategy for the task, for the given domain, extracted actions, described agents, and planning goal.
\end{enumerate}
Therefore, the VLM returns the initial configuration of the agents and obstacles in the video frame, and an ideal strategy to pursue the planning goal.

\subsubsection{Role Retrieval.} A description of the relevant roles in the multi-agent team is provided to the Coach VLM in the input prompt. In this way, the Coach module implicitly identifies the roles of the robots in the multi-agent strategy, by mapping the robot in the video frame to the role assigned in the high-level plan, based on its relevance in the strategy. 
To enable the coach to do so, relevant roles, such as the goalie (the goalkeeper), the jolly (tasked with moving around to receive passes), or the striker (tasked with managing the ball) are defined in natural language in the input prompt. For example, the striker role is described as follows:
\begin{small}
\begin{verbatim}
- STRIKER: The striker robot is responsible for scoring goals, and is 
allowed to perform the following actions: pass_the_ball and kick_to_goal
...
\end{verbatim}
\end{small}
Similarly, it is possible to describe waypoints (locations reachable by robots on the field) in natural language:
\begin{small}
\begin{verbatim}
OUR_GOAL: Our team's goal area.
...
\end{verbatim}
\end{small}
Having discretized the field in a distribution of waypoints, it is possible to identify robot roles by comparing their position on the field with the assignment to waypoints that the VLM returns in its first task. 

\subsection{Plan Refinement}
The plan obtained up until now is a high-level abstract plan, containing tokens representing real-world entities, such as agents, the ball, waypoint locations on the field, and actions. Still, to be executed, two problems have to be solved:
\begin{itemize}
    \item The plan must be translated into a more structured version, so that a parser may be used to build a Finite-State Machine that is readily executable by the robotic agents.
    \item Given that several agents concur with the execution of this plan, a hint on the synchronization between the actions of different agents must be provided.
\end{itemize}

\subsubsection{Plan Grounding.} The first issue can be solved by transforming the original high-level abstract plan, into a more structured version of the plan. To this aim, the LLM is instructed to satisfy the following task:
\begin{small}
\begin{verbatim}
Generate a high-level plan for the team using the coach's recommendations 
in the shortest number of steps using the planning domain, the planning 
goal, and the scenario. Don't write actions for the opponent team players.
\end{verbatim}

\end{small}
This simple task prompt is then followed by:
\begin{itemize}
    \item The same planning domain provided to the VLM.
    \item The actions retrieved by the action database.
    \item Constraints about the structure of the plan, to make sure that actions are written as \verb|ACTION_ID AGENT_ID ARGUMENTS| (where \verb|AGENT_ID| is the agent acting), followed by a list of possible values for the \verb|ACTION_ID| token (the retrieved actions).
    \item A list of possible values for the \verb|AGENT_ID| field, containing all the relevant roles for the plan, and descriptions of the agents, in natural language.
    \item Several more fine-grained constraints to ensure that the result is desirable.
    \item The "scenario", is the configuration of elements on the field extracted from the output of the VLM.
    \item Finally, the high-level plan to be refined, presented as "COACH RECOMMENDATIONS"
\end{itemize}
The result is a more structured version of the high-level plan returned by the "coach" VLM, with action names and arguments correctly grounded in the domain tokens.

\subsubsection{Plan Synchronizer.} At this stage, actions in the generated plan feature multiple agents but still follow a sequential order. To obtain a multi-agent plan, there should be some indication of how actions performed by different agents at the same time should be synchronized. For this reason, the component of \textit{Plan Parallelization} instructs the LLM to return a plan where actions that are executed by different agents at the same time are put together in a "\verb|JOIN{...}|" block. The prompt provided to the LLM is then further enriched with a positive example of a valid result and several negative examples of results to avoid, with a brief explanation of the reason that makes them invalid. The resulting plan will then look like:
\begin{verbatim}
pass_the_ball STRIKER {'SENDER': STRIKER, 'RECEIVER': JOLLY}
JOIN {pass_the_ball JOLLY {'SENDER': STRIKER, 'RECEIVER': JOLLY},  
      kick_to_goal JOLLY {}}
\end{verbatim}

\subsection{Plan Execution}
The offline portion of the pipeline is used to generate a collection of plans from a given set of video frames, as in Fig. \ref{fig:pipeline}.
Collected plans are then selected based on their similarity to the real-world scenario.
The similarity between the two scenarios can be performed using a clustering technique where the current game scenario is obtained by the global model perceived by the robot.
An accurate generation of plans based on various game situations allows for the handling of almost all game situations during matches and allows the robots to apply when needed a suitable behavior that maximizes the likelihood of scoring a goal.

\section{Experimental Results}
\label{sec:ExperimentalResults}

This section illustrates the experimental results obtained from the evaluation in a simulated environment of the results of the current work, with the primary objective of assessing its effectiveness in generating high-quality plans collected by processing video frames from real RoboCup SPL matches.  

\subsubsection{Prompt Engineering.} To obtain the desired results, an initial phase of prompt refinement was required. The prompt of the VLM "coach" module could be more information-rich with a smaller risk of hallucination, given the multi-modal input. Providing both a video frame and textual input allows the coach to give better insight into the subsequent stages along the pipeline. The accuracy of the output at this stage is key, as any error at this stage could mislead the entire generation pipeline. The application domain constitutes a critical component of the prompt for this module, particularly in a use case where the objective is to describe the location of the players based on both an image and a text description.
The main obstacle is the representation of 2D coordinates within the provided frame, which cannot be easily achieved using a VLM due to its difficulty in working on cartesian coordinates.
Instead of coordinates, tokens representing waypoints are used as a way of discretizing the field, by marking significant field locations. The description of these tokens should not mislead the VLM and the significance of waypoints must be explained in natural language. For example, an earlier iteration of the prompt contained the waypoint "KICKING\_POSITION": initially described as a "vantage point", whose semantics are far from having a geometric value. The final version of the prompt avoids such bias-inducing mistakes by describing it as a "favorable position". This improved dramatically the results of the scenario description task.

The second task assigned to the coach consisted of generating a plan given the situation depicted in the provided input frame. Ideally, the plan should be structured according to an easily parsable template. A first attempt at generating a structured plan directly using the VLM resulted in heavy hallucination. To overcome this problem, the VLM was instead tasked with generating a high-level plan, as if it were a soccer coach. To exert more control on the overall strategy, the possibility to customize the overall "tactics" was added, for example by specifying whether passes were desirable or if the strategy should be offensive or defensive. A subsequent plan refinement module was added, to refine the high-level plan into a parsable plan. This module used only a textual input, therefore it was key to minimize the input token count, to avoid hallucination. 

\subsubsection{Setup.} The VLM Coach module relies on the OpenAI \emph{GPT 4 Turbo} model with vision while the text-only modules feature the latest OpenAI \emph{GPT 3.5 Turbo} model \emph{gpt-3.5-turbo-0125}. Tests were conducted in a simulated environment based on SimRobot \cite{10.1007/11780519_16}. Pertinent frames were extracted from historical SPL matches, publicly accessible on the web. For every frame, the robot configuration was replicated within the simulation, reflecting each robot's pose $p_i = (x,y,\theta)$ as extracted from the MARIO, ensuring a faithful recreation of real-world scenarios within the simulated environment. Starting from the same initial configuration, the evolution of plans was considered, comparing the pre-existing behaviors and the plans generated by the presented pipeline. For simplicity, only offensive scenarios were taken into account due to their complexity and larger action space compared to defensive scenarios.

\subsubsection{Evaluation.} The proposed approach is evaluated using the success rate metric, measuring the team's ability to score goals in offensive situations during gameplay. To assess the effectiveness of our method, we compared two approaches: the baseline approach, relying on the predefined behaviors of the SPQR Team 2023 code release, and the dynamic plans generated by the Large Language Model (LLCoach). In addition to the success rate, we considered the average number of passes and average scoring time in seconds as supplementary metrics to illustrate the evolution of gameplay dynamics over different settings.
Considering two different situations from the final match BHuman - HTWK we recreated the robot configurations running eight simulations in total. It is worth noting that, since several API calls are required to obtain and refine the plan throughout the multi-role pipeline, the overall execution time is affected mainly by the inference times of LLM/VLM agents.
 
\begin{table}[t]
\centering
\begin{tabular}{|c|c|c|}
\hline \textbf{Metrics}
 & \textbf{SPQR Human-written Code} & \textbf{LLCoach} \\
\hline
Success Rate & 30\% & \textbf{90\%} \\
\hline
Avg. no. of passes & 0.60 & \textbf{1.33} \\
\hline
Avg. scoring time & 66 sec. & \textbf{29.7 sec.} \\
\hline
\end{tabular}
\caption{Comparison of success rate, average passes, and seconds played obtained by comparing eight simulations with LLCoach and the human-written code.}
\vspace{-0.7cm}
\end{table}

\section{Conclusions and Future Directions}
\label{sec:conclusion}
In this paper, we presented a novel approach to plan generation, exploiting the common knowledge in Large Language Models. The proposed approach represents a first attempt at using generative AI in the context of robot soccer. In particular, we described a complete pipeline, in four stages, where a high-level plan, is progressively refined into a multi-agent plan. The first stage, a Visual Language Model acting as a coach, generates plans from video frames, extracted from videos RoboCup SPL matches. Collected plans can then be run by real NAO robots, as shown by the promising results conducted in the context of RoboCup SPL matches.
As future work, we intend to use videos from human soccer matches to create policies for soccer robot players. 
Our aim is to reduce the gap between robot behaviors and human moves in order to achieve the 2050 RoboCup goal.

\section{Acknowledgement}
This work has been carried out while Michele Brienza was enrolled in the Italian National Doctorate on Artificial Intelligence run by Sapienza University of Rome. This work has been supported by PNRR MUR project PE0000013-FAIR.

%
%
%
\bibliographystyle{splncs04}
\bibliography{biblio}

\end{document}